\documentclass[11pt]{article}

\usepackage[preprint]{acl}

\usepackage{times}
\usepackage{latexsym}
\usepackage{booktabs}
\usepackage[T1]{fontenc}

\usepackage{makecell}

\usepackage{microtype}

\usepackage{inconsolata}

\usepackage{graphicx}

\usepackage{tikz}
\newcommand{\highlight}[1]{\tikz[baseline=(X.base)] \node[rectangle, fill=green!15, inner sep=2pt] (X) {#1};}
\newcommand{\highlightred}[1]{\tikz[baseline=(X.base)] \node[rectangle, fill=red!15, inner sep=2pt] (X) {#1};}

\title{SI-Bench: Benchmarking Social Intelligence of Large Language Models in Human-to-Human Conversations}

\author{
Shuai Huang \quad Wenxuan Zhao \quad Jun Gao \\
Hello Group \\
\texttt{\{huang.shuai, zhao.wenxuan, gao.jun\}@hellogroup.com}
}

\begin{document}
\maketitle
\begin{abstract}
As large language models (LLMs) develop anthropomorphic abilities, they are increasingly being deployed as autonomous agents to interact with humans. However, evaluating their performance in realistic and complex social interactions remains a significant challenge. Most previous research built datasets through simulated agent-to-agent interactions, which fails to capture the authentic linguistic styles and relational dynamics found in real human conversations. To address this gap, we introduce \textbf{SI-Bench}, a novel benchmark designed to evaluate aspects of social intelligence in LLMs. Grounded in broad social science theories, SI-Bench contains 2,221 authentic multi-turn dialogues collected from a social networking application. We further selected a subset of 312 dialogues for manual annotation across 8 major models. The experiments show that SOTA models have surpassed the human expert in process reasoning under complex social situations, yet they still fall behind humans in reply quality. Moreover, introducing Chain-of-Thought (CoT) reasoning may degrade the performance of LLMs in social dialogue tasks. All datasets are openly available at \href{https://github.com/SI-Bench/SI-Bench.git}{https://github.com/SI-Bench/SI-Bench.git}.
\end{abstract}

\section{Introduction}
Social intelligence is the ability to understand others and act wisely in human relations \cite{thorndike1920intelligence}. In recent literature, social intelligence is understood as comprising social awareness and social facility \cite{goleman2006social}. This capacity enables individuals to get along well with others and secure their cooperation \cite{albrecht2005social}, giving it clear practical value in everyday life. With the fast development of AI, social intelligence is regarded as foundational to successful human-agent interaction and collaboration \cite{williams2022supporting}. 

\begin{figure}[h]
  \includegraphics[width=\columnwidth]{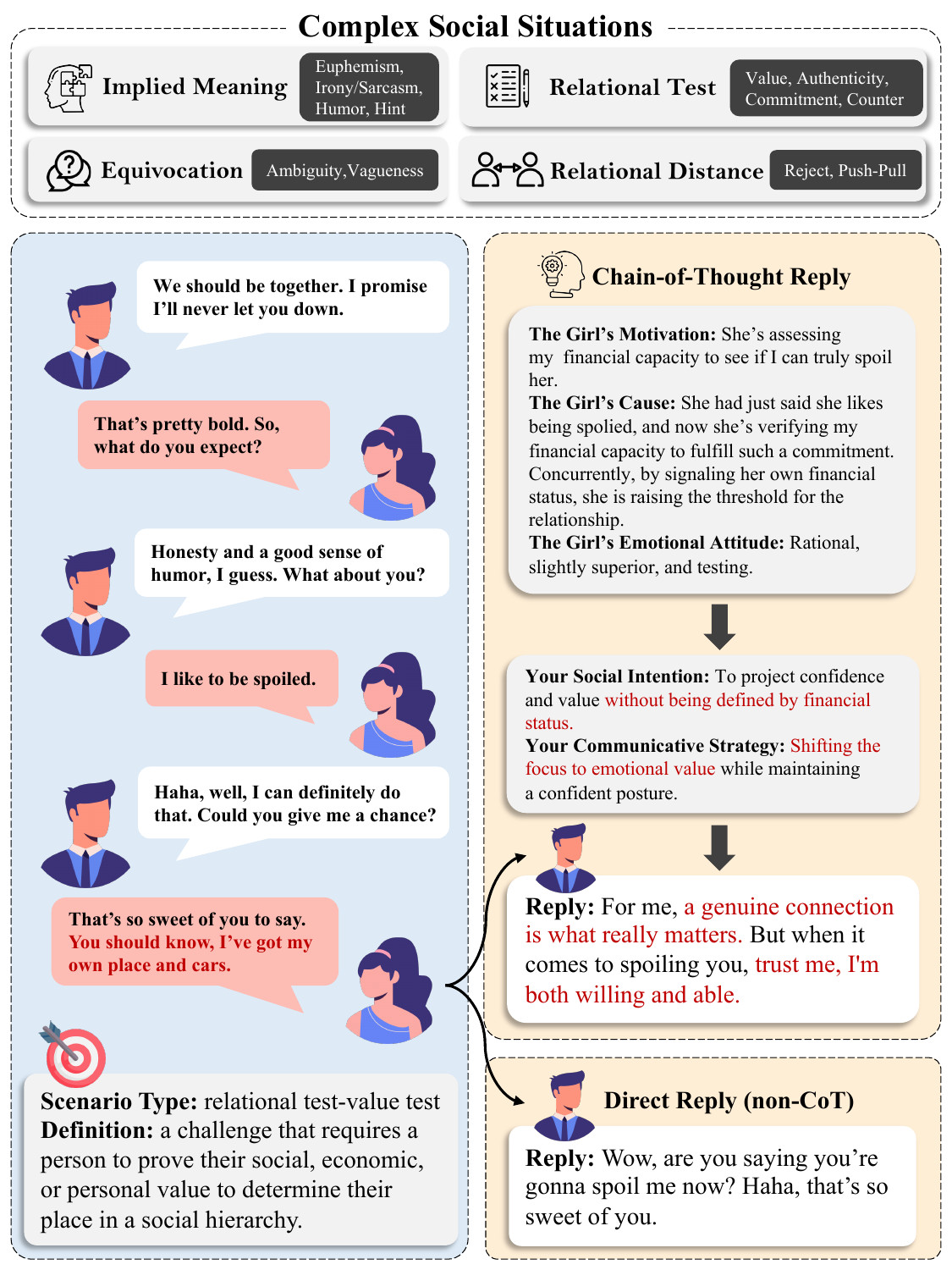}
  \caption{Overview of the SI-Bench framework and a sample model response.}
  \label{fig:SI-Bench}
\end{figure}

Although the importance of social intelligence is recognized, methods for evaluating this capacity in Large Language Models (LLMs) remain very limited. Current evaluations face two key limitations: (1) lack of ecological validity: many benchmarks rely on static texts scenarios in formats like multiple-choice questions \cite{sap2019socialiqacommonsensereasoningsocial} or short-text judgments \cite{forbes2021socialchemistry101learning}. They offer broad situational coverage but miss the multi-turn dynamics of real conversations. Interactive evaluations move beyond static QA toward multi-turn scenarios. Benchmarks such as SOTOPIA \cite{zhou2024sotopiainteractiveevaluationsocial}, AgentSense \cite{mou2024agentsensebenchmarkingsocialintelligence}, EgoSocialArena \cite{hou2025egosocialarenabenchmarkingsocialintelligence}, and SocialEval \cite{zhou-etal-2025-socialeval} advance this paradigm by involving multi-agent simulations and action-level feedback. However, their reliance on scripted settings leaves a gap with the dynamic strategies of authentic human social interaction. (2) conflated evaluation dimensions: existing evaluations \citep{he2023hitombenchmarkevaluatinghigherorder,chen2024tombenchbenchmarkingtheorymind} primarily score the final output, treating the model's internal reasoning process as an unobservable black box. This makes it difficult to diagnose the root cause of failures: does the model fail because it misunderstands the social context, or because it cannot translate a correct understanding into an appropriate response? While recent work has begun to assess process capabilities alongside goal achievement \cite{zhou-etal-2025-socialeval}, the field is still converging on a standard methodology.

To fill the gaps, we introduce \textbf{SI-Bench}, a new benchmark of social intelligence in LLMs. Built upon classic social science theories, SI-Bench covers 12 representative types of complex social situations, aiming to capture the real-world reasoning patterns and communication strategies humans employ when facing social challenges. SI-Bench contains 2,221 authentic multi-turn dialogues collected from a social networking application. These dialogues include numerous Chinese dialects, slang, and colloquial expressions, reflecting rich linguistic diversity and the authentic distribution of social pragmatics. In each dialogue, the final utterance corresponds to one of the 12 defined social situations, and the LLMs act as the user's proxy, generating a response to that message based on the conversational context. On this basis, we selected 312 dialogues and manually annotated the replies of 8 leading LLMs, using a human expert's handwritten replies as the baseline for comparison. By introducing the Chain-of-Thought (CoT) reasoning process for social situations, we independently evaluate both process quality and outcome quality. The evaluation dimensions include: (1) process quality, covering five aspects, including motivation, reason, emotional attitude, social intent, and communication strategy; (2) final reply quality, assessed on linguistic appropriateness, relational progression, and strategic sophistication. Moreover, through pairwise human annotations of wins and losses, we compare the response quality of CoT-guided replies with direct (non-CoT) replies, aiming to reveal the underlying mechanisms of CoT in social dialogue tasks. A complete overview of the situation taxonomy and model response examples is illustrated in Figure~\ref{fig:SI-Bench}.

Our contributions are as follows:
\begin{itemize}
    \item We introduce and release SI-Bench, to our knowledge, the first benchmark built on real human dialogues for evaluating social intelligence in LLMs. It covers diverse and challenging social situations, enabling a comprehensive evaluation of LLMs' social intelligence.

    \item We design an evaluation framework that decouples the model's reasoning process from its reply quality. This enables a fine-grained attribution of model performance, providing directions for future research.

    \item Our experiments reveal a significant thought-action gap in LLMs. Some leading models outperform the human expert in process quality scores, yet they still fall behind human in final reply quality. We also show that introducing CoT can be harmful for reply generation.
\end{itemize}

\section{Benchmark Construction}

\subsection{Data Source}

Our evaluation dataset is derived from authentic dyadic conversations on a leading social networking platform in China, with all data carefully anonymized and filtered. Compared with existing datasets that rely on human-authored scripted  scenarios \cite{sap2019socialiqacommonsensereasoningsocial,zhou-etal-2025-socialeval}, public forum interactions \cite{xu2024academicallyintelligentllmsnecessarily}, or open source movie scripts \cite{mou2024agentsensebenchmarkingsocialintelligence}, our dataset offers significantly higher ecological validity \cite{bronfenbrenner1977toward}. These conversations occur in authentic social interaction scenarios, where the main objective is to build deeper interpersonal relationships. Such contexts are inherently characterized by probing, negotiation, and subtle manipulation. Our focus is not on casual small talk, but rather on complex social situations. These are instances where one party poses statements with implicit challenges or high uncertainty, making the other party's response particularly difficult.  Inappropriate replies at these junctures may lead to conversation termination or relationship deterioration. These high risk and high difficulty critical moments most effectively test LLMs' social strategy selection and adaptive response capabilities.

\subsection{A Taxonomy of Complex Social Situations}
To systematically analyze the complex social situations within dyadic conversations , we propose a taxonomy grounded in classic social science theories. This framework is structured around two fundamental tensions inherent in social interaction: \textbf{semantic uncertainty} and \textbf{power dynamics}. 

\paragraph{Tension I: Communication Challenges under Semantic Uncertainty}
Semantic uncertainty occurs when individuals with asymmetric information interpret ambiguous signals differently, consistent with Hall's description of a high-context communication \cite{hall1976beyond}. For our analysis, we divide this uncertainty into two types. 

\textbf{Implied meaning} According to Grice \cite{grice1975logic}, when the literal meaning of an utterance diverges from the speaker's true intention, the listener must rely on shared contextual knowledge to infer the implicit meaning. We classify this phenomenon in social interaction into four types: (a) \textbf{Euphemism}: the strategic use of indirect language to avoid potential social conflict, a behavior rooted in the need to preserve face \cite{goffman1967interaction_anchor}; (b) \textbf{Irony / Sarcasm}: the use of opposing literal and intended meanings to convey criticism, requiring shared context for interpretation \cite{kreuz1989sarcastic}; (c) \textbf{Hint}: the speaker leaves things unsaid and provides only partial information, prompting the listener to fill in the missing meaning from context and grasp the intended message \citep{brown1987politeness,carston2002thoughts}; (d) \textbf{Humor}: a strategic communicative tool for building rapport or drawing social boundaries \cite{meyer2000humor}. This scenario requires LLMs to engage in pragmatic reasoning that goes beyond literal comprehension to accurately infer the speaker's underlying purpose.

\textbf{Equivocation} Equivocation theory posits that when a speaker faces a communicative situation where any direct response will lead to negative consequences. Referred to as an avoidance-avoidance conflict, this motivates the speaker to use ambiguous messages to avoid committing to a clear position \cite{bavelas1990equivocal}. This uncertainty takes two forms: (a) \textbf{Ambiguity}: the strategic use of unclear expressions that allow multiple interpretations \cite{Eisenberg01091984}; (b) \textbf{Vagueness}: the use of intrinsically uncertain expressions that serve specific conversational goals \cite{channell1985vagueness}. In contrast to implied meaning, where the unstated intent is definite, the intent behind an equivocation expression is uncertain. Therefore, the LLMs' key task is not passive guessing but active clarification to resolve ambiguity.

\paragraph{Tension II: Social Challenges under Power Dynamics}
Social interaction often revolves around the negotiation of relational dominance. In such cases, communication frequently takes a low-context form, as described by Hall \citeyearpar{hall1976beyond}. We divide these cases into two categories.

\textbf{Relational Distance} This category includes two communicative behaviors aimed at managing psychological distance. (a) \textbf{Rejection}: the explicit or implicit refusal of a request or proposal. In Brown and Levinson's \citeyearpar{brown1987politeness} politeness theory, an act of rejection is considered a face-threatening act. It directly threatens the recipient's positive face, which is the need for belonging and approval. (b) \textbf{Push-Pull Dynamics}: a pattern of suddenly reducing responsiveness after a period of high engagement. This pattern mirrors demand/withdraw interactional cycle in marital conflict, where one party retreats via defensiveness and passive inaction in response to demands for change \cite{christensen1990gender}. This scenario tests LLMs' ability to perform relationship repair and maintain the conversation when encountering social setbacks.

\textbf{Relational Test} In social interactions, people use various communication strategies to test or assess others, helping them make better relationship decisions. (a) \textbf{Value Test}: a request that the other party demonstrate their worth, consistent with principles of social exchange theory \cite{homans1958social}; (b) \textbf{Authenticity Test}: an act of expressing explicit doubt or challenge toward the authenticity of another's self-presentation, such as their appearance, status, or stated experiences. Following Goffman's \citeyearpar{goffman1959presentation} dramaturgical approach, individuals engage in performances that may be accepted or challenged by their audience; (c) \textbf{Commitment Test}: a question or statement that measures the other party's sincerity and exclusivity in current relationship, relative to other potential partners. This concept aligns with specific secret tests identified by Baxter and Wilmot \citeyearpar{baxter1984secret};  (d) \textbf{Counter Test}: a response to a perceived challenge by reframing the topic, questioning the challenger's legitimacy, or initiating a counter challenge. This represents a proactive move within a frame dispute \cite{goffman1974frame}. This scenario tests whether a model can show relationship management capabilities.

\begin{figure*}[!htbp]
  \includegraphics[width=\textwidth]{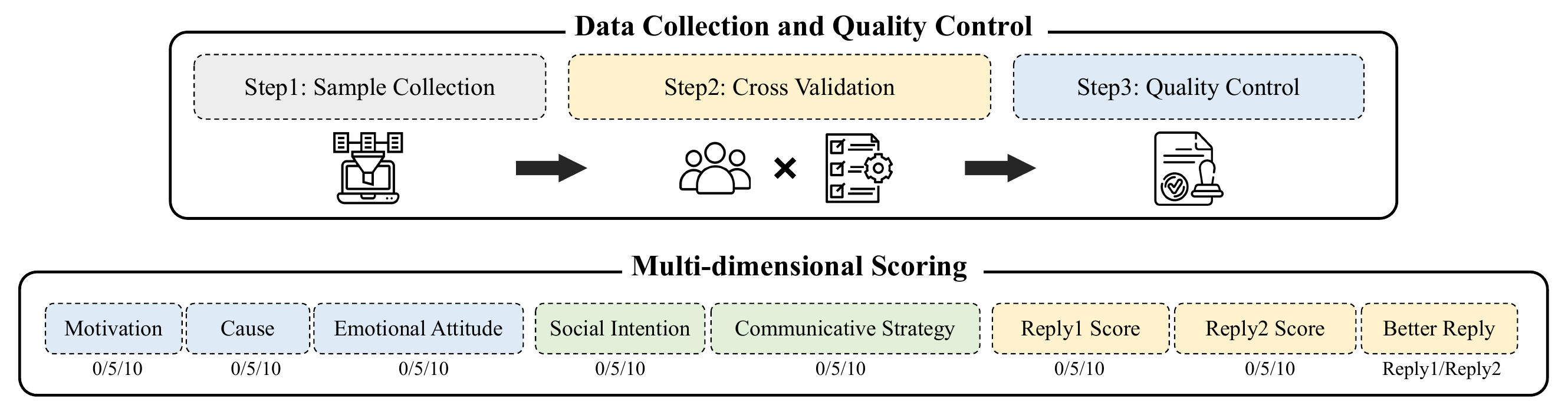}
  \caption{Data Collection and Scoring Pipeline.}
  \label{fig:sample_selection}
\end{figure*}

\subsection{Data Collection}

We design a three-step sample construction pipeline, as shown in Figure~\ref{fig:sample_selection}, which includes sample collection, cross validation and quality control procedures.

\textbf{Sample Collection} We first extract multi-turn dialogues collected from a social networking application where the other party's utterance ends the conversation and poses potential complex situations or social challenges to our side as candidate samples. 

\textbf{Cross Validation} We organize four annotators into two pairs to perform cross validation on the candidate samples. Annotator A labels and justifies samples meeting predefined criteria, and annotator B independently reviews them. Only samples with complete agreement are kept to ensure reliability.

\textbf{Quality Control} To protect privacy and enhance data quality, we perform minimal rewriting on verified samples. We remove identifiable information, shortened overly long dialogues while keeping key context, and ensure each sample retain at least six turns.

\section{Evaluation Framework}
We propose a multi-dimensional evaluation framework that goes beyond traditional outcome-based scoring. Inspired by social science theories, it evaluates both the model's reasoning process and its final replies within social interactions.

\subsection{Theoretical Foundation}
A successful response involves a complete chain of perception, planning, and execution. To conduct fine-grained evaluation of this process, we draw inspiration from the classic social information processing theory \cite{crick1994review}. This theory decomposes an individual's response to social events into a series of cognitive steps. Following this framework, we divide the model's CoT output and final reply into three stages: 

\subsubsection{Contextual Understanding} Corresponding to the ``encoding of cues'' and ``interpretation of cues'' stages in the theory. This stage evaluates the model's ability to accurately understand the underlying motivations, causes, and emotional attitudes behind the other party's utterances from their perspective. We model this process as a perception function, $f_{P}$:
\begin{equation}
    S_{o} = f_{P}(C)
\end{equation}
where $C$ represents the complete dialogue \textbf{C}ontext. $S_{o}$ represents the model's inferred mental \textbf{S}tate of the \textbf{o}ther party. This state is a tuple of three elements:
\begin{equation}
    f_{P}(C) = \langle m, c, e \rangle
\end{equation}
These elements directly correspond to our evaluation dimensions: \textbf{m}otivation, \textbf{c}ause, and \textbf{e}motional attitude.

\subsubsection{Response Strategy} Depending on the ``clarification of goals'',  ``response access or construction'', and ``response decision'' stages in the theory, the model adopts the speaker's perspective to achieve goals based on its understanding, including establishing social intentions and selecting communicative strategies. We model this process as $f_{S}$:
\begin{equation}
    R_{m} = f_{S}(S_{o})
\end{equation}
where $R_{m}$ represents the \textbf{R}esponse strategy of the \textbf{m}odel. Based on our evaluation dimensions, this state is a tuple of two elements:
\begin{equation}
    f_{S}(S_{o}) = \langle i, s \rangle
\end{equation}
These elements directly correspond to our evaluation dimensions: social \textbf{i}ntention and communicative \textbf{s}trategy.

\subsubsection{Reply Generation} 
Align with the ``behavioral enactment'' stage of the theory, this stage evaluates the model's ability to generate the final reply. We model this process as a \textbf{g}eneration function $f_{G}$:
\begin{equation}
    U_{m} = f_{G}(S_{o},R_{m})
\end{equation}
where $U_{m}$ is the final \textbf{u}tterance generated by the \textbf{m}odel.

\subsubsection{Overall Assessment}
The general social intelligence of the model is evaluated based on two components: the quality of its reasoning process, $Q_{proc}$, and the quality of its final reply, $Q_{rep}$. The process quality $Q_{proc}$ is defined as follows:
\begin{equation}
    Q_{proc} = E(S_{o},R_{m})
\end{equation}
The reply quality $Q_{rep}$ is formulated as follows:
\begin{equation}
    Q_{rep} = E(U_{m})
\end{equation}
where ${E}$ represents the evaluation function.

\subsection{Evaluation Rubric}
To ensure consistency, we design a detailed three-tier scoring system (0–5–10) for both process and outcome evaluations. 

\textbf{Process Evaluation}
This evaluation assesses the depth and breadth of the model's CoT output. A simplified scoring criteria are as follows. For details, see Table~\ref{tab:process_criteria}.
\begin{itemize}
    \item 0: The analysis contains errors.
    \item 5: The analysis is largely correct but superficial.
    \item 10: The analysis is accurate and demonstrates deep insight.
\end{itemize}

\textbf{Reply Evaluation}
The outcome evaluation measures the quality and effectiveness of the model's final reply within a social interaction. According to relational control theory \cite{rogers1975analysis}, every message exchange not only conveys content but also defines and negotiates the relationship and power dynamics between parties. In our work, the other party constructs complex situations that position the speaker in a passive position. Therefore, a high-quality response is defined by its ability to effectively manage these relational dynamics. We provide the simplified scoring criteria below. The complete version appears in Table~\ref{tab:evaluation_criteria}. 
\begin{itemize}
    \item 0: The response has significant flaws.
    \item 5: The response is adequate, but tends toward passive defense.
    \item 10: The response is excellent and proactively takes control of the dialogue.
\end{itemize}

\subsection{Evaluation Process}
\begin{figure}[h]
  \includegraphics[width=\columnwidth]{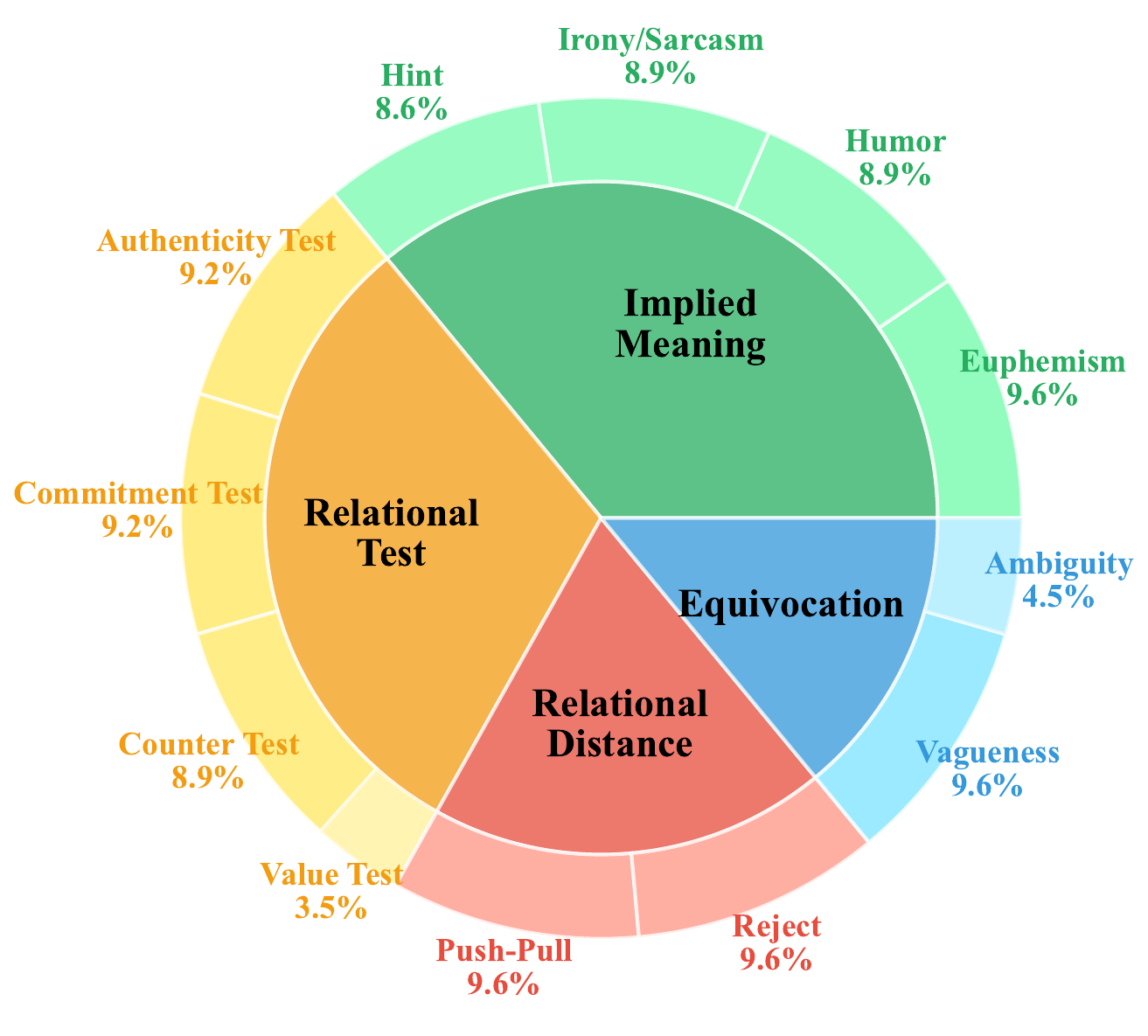}
  \caption{Distribution of Complex Situations in the Experimental Dataset.}
  \label{fig:dataset_distribution}
\end{figure}

We sample 312 dialogues for independent annotation. The situation distribution in our experimental dataset is balanced, as shown in Figure~\ref{fig:dataset_distribution}. We recruit 3 graduate students with backgrounds in psychology and linguistics as annotators. For each sample and evaluated model, annotators score the model's CoT reasoning process (Motivation, Cause, Emotional Attitude, Social Intention, Communicative Strategy) and two anonymized final replies (Reply CoT and Reply Direct, randomly labeled as ``Reply 1'' and ``Reply 2'') to avoid bias. A forced preference judgment is applied when scores are identical, requiring annotators to choose which reply is better. To handle the subjectivity in scoring, we aggregate the 3 annotators' ratings using the arithmetic mean instead of majority voting. The agreement between human annotators is acceptable, with a Krippendorff's Alpha score of 0.712 \cite{krippendorff}.

\section{Experiment}
We evaluate 8 advanced LLMs including Claude-4-Opus, Claude-4-Sonnet, GPT-4o, Gemini-2.5-Pro, Gemini-2.5-Flash, Doubao-1.6-Thinking, Doubao-1.5-Pro-Character, and Qwen-2.5-Max. We recruit a graduate student with a background in psychology as the human expert, who manually authors responses for each sample following the same evaluation dimensions used for the LLMs. Our experiments aim to answer the following research questions:

\begin{table*}[h]
\centering
\footnotesize
\setlength{\tabcolsep}{2pt}
\begin{tabular}{l|ccc|cc|cc|c}
\toprule
\textbf{Model} & \textbf{Motivation} & \textbf{Cause} & \textbf{\makecell{Emotional \\Attitude}} & \textbf{\makecell{Social \\Intention}} & \textbf{\makecell{Communicative \\Strategy}} & \textbf{CoT Reply} & \textbf{Direct Reply} & \textbf{CoT Win Rate} \\
\midrule
Claude-4-Opus & 7.62 & 7.57 & 8.95 & 7.47 & 6.80 & 5.32 & 5.28 & \highlightred{37.8\%} \\
\midrule
Claude-4-Sonnet & 7.50 & 7.28 & 8.90 & 7.85 & 7.37 & 5.07 & 5.37 & \highlightred{31.2\%} \\
\midrule
Doubao-1.5-Pro-Character & 2.63 & 2.63 & 4.61 & 6.51 & 6.25 & 4.63 & 5.00 & \highlightred{33.1\%} \\
\midrule
Doubao-Seed-1.6 & 5.97 & 6.05 & 7.24 & 6.54 & 6.28 & 5.06 & 5.47 & \highlightred{32.9\%} \\
\midrule
Qwen-2.5-Max & 5.20 & 5.15 & 7.15 & 7.35 & \highlight{\textbf{7.69}} & 4.85 & 4.47 & \highlightred{39.0\%} \\
\midrule
GPT-4o & 5.63 & 5.72 & 7.65 & 7.03 & 7.06 & 5.07 & 4.54 & \highlightred{48.6\%} \\
\midrule
Gemini-2.5-Flash & 5.66 & 6.49 & 7.93 & 7.57 & 7.01 & 4.56 & 4.61 & \highlightred{33.2\%} \\
\midrule
Gemini-2.5-Pro & \highlight{\textbf{8.04}} & \highlight{\textbf{8.01}} & \highlight{\textbf{9.06}} & \highlight{\textbf{8.22}} & 6.99 & 5.73 & 5.94 & \highlightred{37.6\%} \\
\midrule
Human & 5.68 & 5.41 & 6.99 & 7.68 & 7.11 & \highlight{\textbf{6.13}} & \highlight{\textbf{6.40}} & \highlightred{43.4\%} \\
\bottomrule
\end{tabular}
\caption[Human and LLMs performance on different dimensions.]{
Human and LLMs performance on different dimensions. \colorbox{green!15}{Green} represents the best-performing models, while \colorbox{red!15}{red} represents a winning rate below 50\%.
}
\label{tab:model_scores}
\end{table*}

\begin{itemize}
\item \textbf{RQ1:} Do even the most advanced models fall behind the human expert in certain dimensions?
\item \textbf{RQ2:} Do models show systematic capability variation across complex social situations?
\item \textbf{RQ3:} How does CoT affect the quality of replies?
\item \textbf{RQ4:} Which CoT dimensions best predict high-quality replies?
\end{itemize}

\subsection{RQ1: Model Performance vs. Human Baseline}

As shown in Table~\ref{tab:model_scores}, models and the human expert show different performance across dimensions.

\textbf{Process Dimensions} SOTA models (Gemini and Claude series)  outperform the human expert, particularly in motivation, cause, and emotional attitude dimensions, with advantages ranging from 1.82 to 2.60 points. This suggests that advanced LLMs have developed capabilities in understanding social contexts and reasoning about the human mind.

\begin{figure}[h]
\includegraphics[width=\columnwidth]{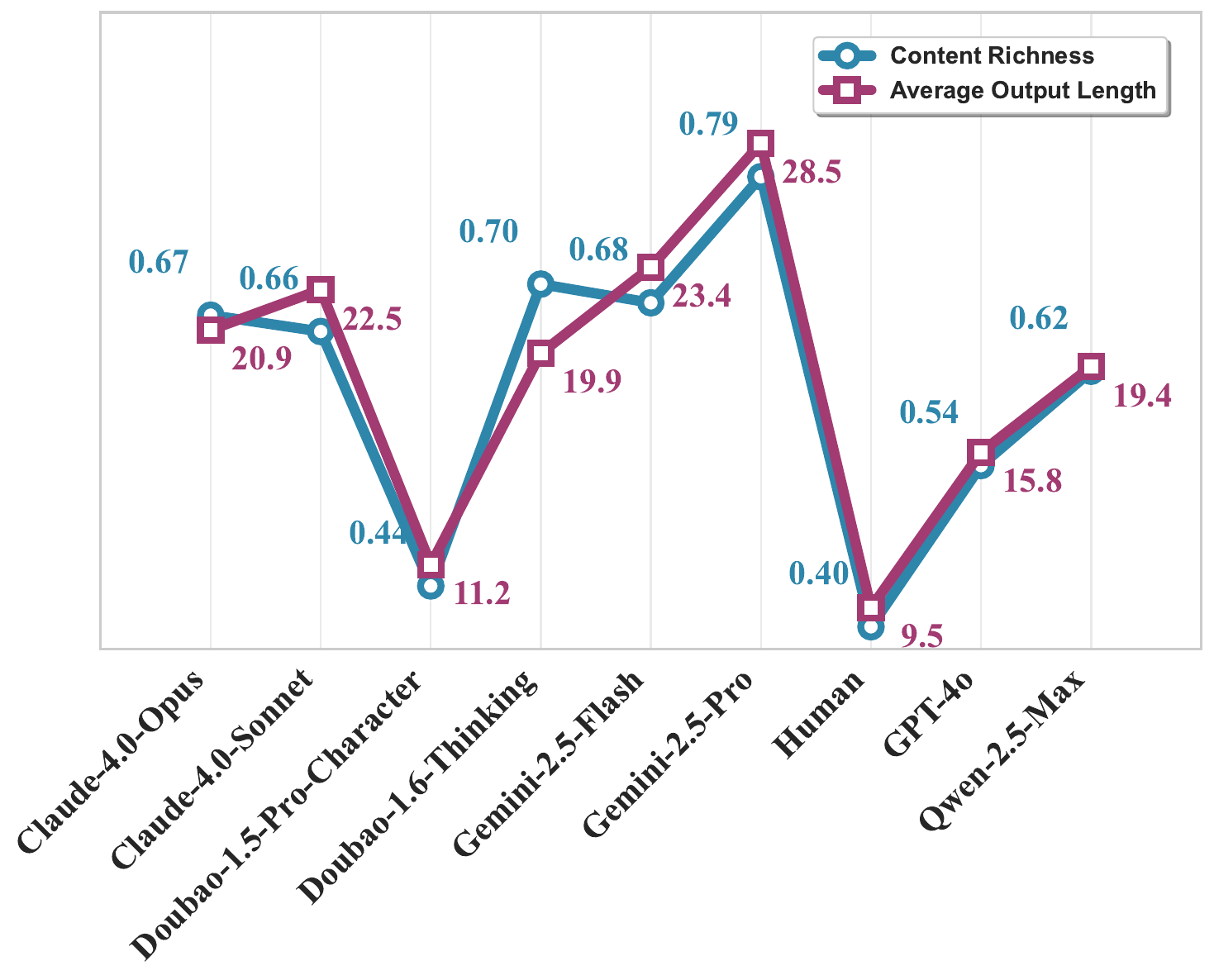}
\caption{Content richness and output length analysis of LLMs and the human expert.}
\label{fig:content_richness_and_length}
\end{figure}

Furthermore, to understand why advanced models outperform the human expert in reasoning process, we conducted content richness (calculated as a weighted score of vocabulary diversity and information entropy) and length analysis comparing LLMs with the human expert, as illustrated in Figures~\ref{fig:content_richness_and_length}. Our analysis reveals that LLMs generate more comprehensive and detailed process, with higher content richness and average word count compared to the human expert. Human often follow their intuition and respond with first-impression answers, while LLMs reason through a wider range of details and possibilities. It enables LLMs to achieve higher scores in process dimensions, where detailed reasoning and multi-perspective analysis are valued.

\textbf{Reply Dimensions} The human expert maintains clear superiority in direct reply quality (6.40 points) , while the best-performing model, Gemini-2.5-Pro, achieves 5.94 points, a gap of 0.46 points. Most other models fall behind human performance by at least 1 point, indicating a real challenge in generating high quality replies.

We observe a noteable phenomenon: the human expert's CoT reply score (6.13 points) is lower than the direct reply score (6.40 points). This result appears counterintuitive, as deliberation is typically assumed to enhance decision quality. We believe this does not indicate a lack of thought from the expert, but is a result of the evaluation method itself. The CoT prompt we design for LLMs is essentially an explicit, sequential scaffold that is intended to simulate the reasoning process. The social reasoning of humans, however, is largely an implicit, parallel, and holistic process that integrates intuition and experience. Forcing the expert to follow this model-centric analytical process interferes with their more advanced cognitive system. This results in CoT replies that are logically correct, but appear socially stiff and unnatural. It reveals the difference between the model's CoT reasoning process and human deep thinking, highlighting the challenge of aligning them.

\subsection{RQ2: Variation in Model Capabilities Across Social Situations}

\begin{figure*}[h]
\includegraphics[width=\textwidth]{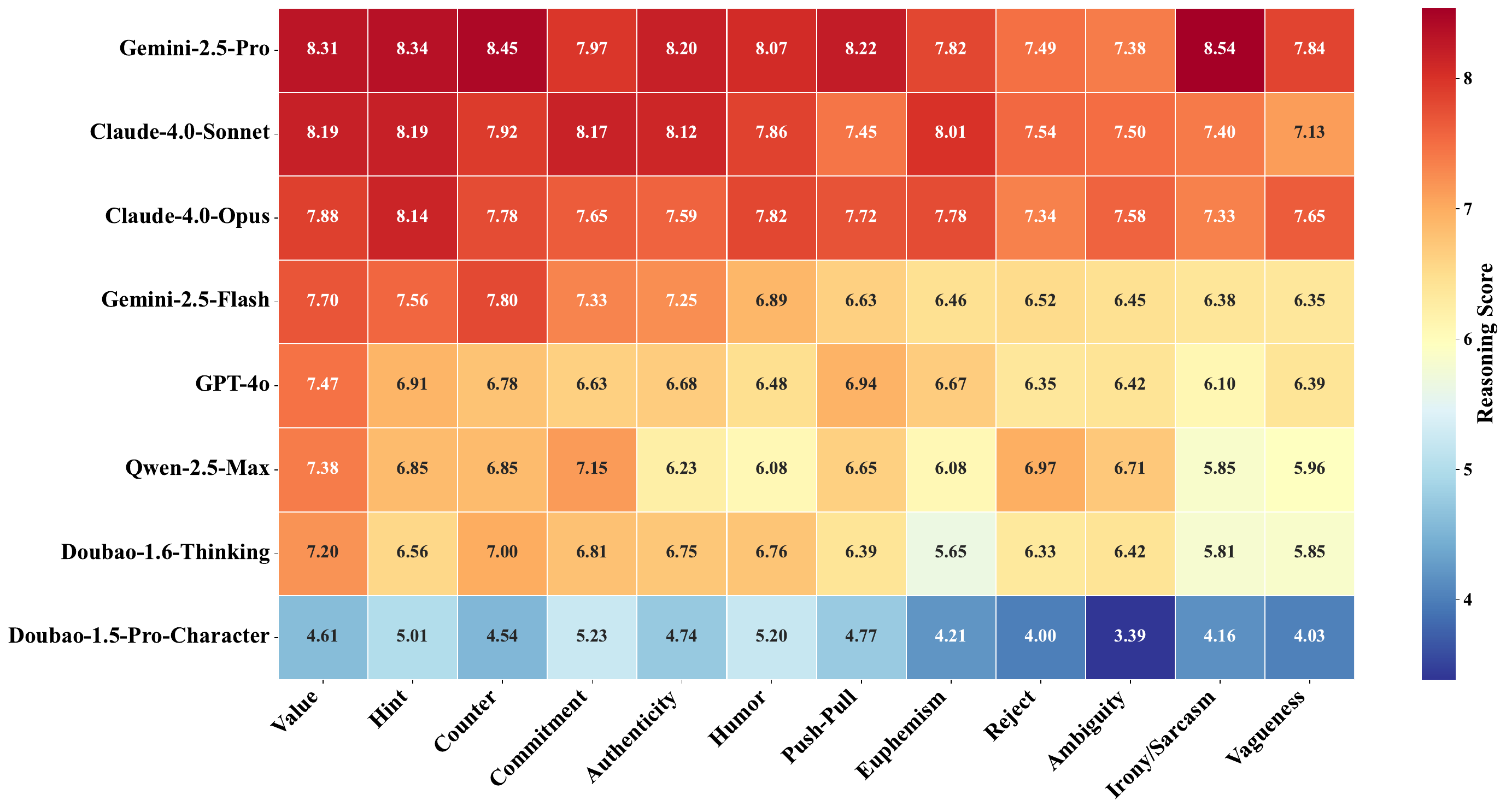}
\caption{Mean scores of reasoning processes across complex social situations.}
\label{fig:scenario_analysis}
\end{figure*}

We analyze the process reasoning performance of different models across 12 complex situations, as shown in Figure~\ref{fig:scenario_analysis}. The results show a clear contextual bias: models generally perform better in low-context situations. Among the top 5 situations with the highest mean reasoning scores, 4 belong to the low-context category (value test, commitment test, counter test, and authenticity test). In contrast, the 4 lowest-scoring situations are all high-context (euphemism, ambiguity, irony/sarcasm, and vagueness). This finding suggests that current models mainly rely on surface semantics for reasoning. They perform well when intentions and goals are explicit, but perform poorly in high-context interactions that require inferring implicit cues. 
\begin{figure}[h]
\includegraphics[width=\columnwidth]{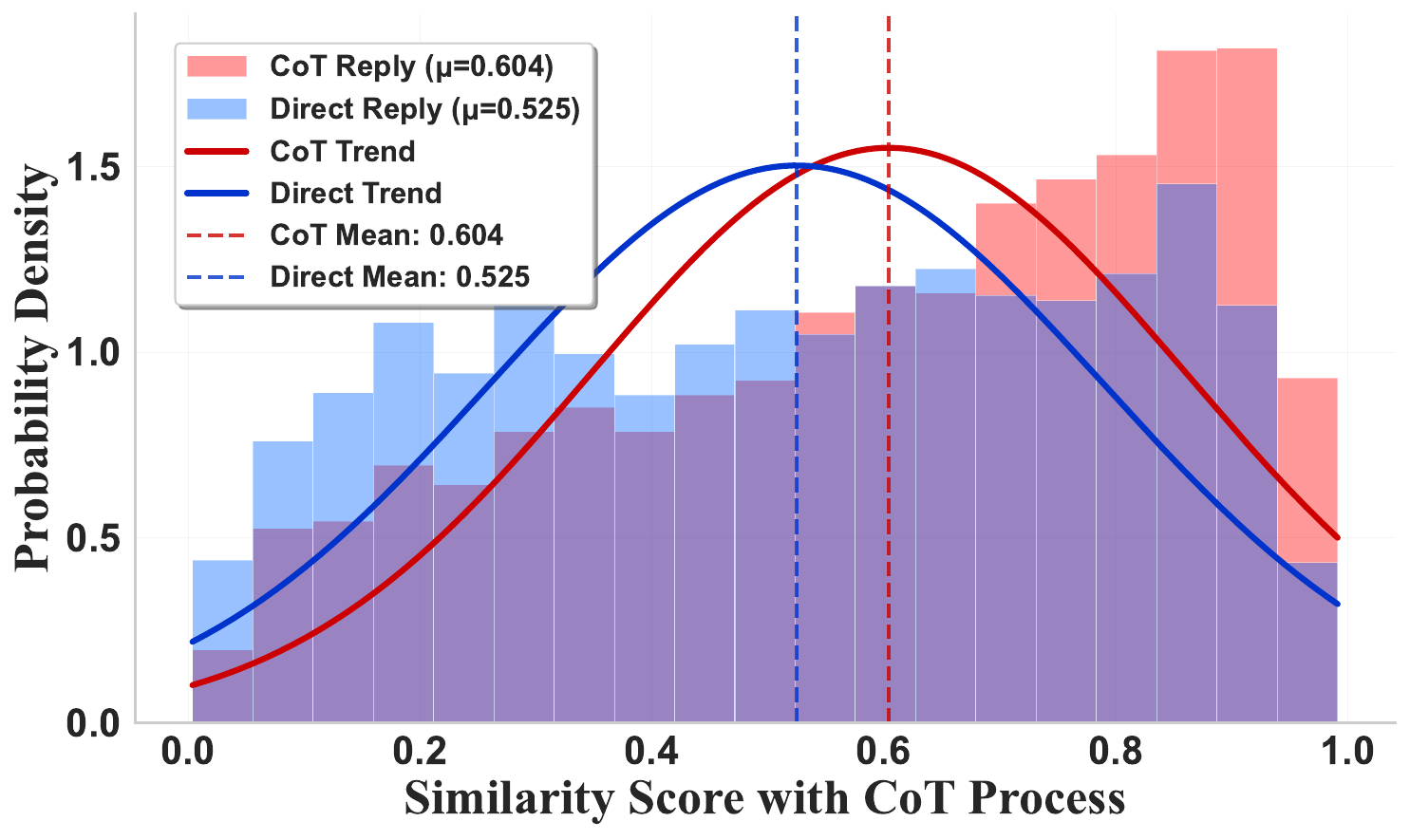}
\caption{CoT reply shows higher similarity with the CoT process than direct reply.}
\label{fig:cot_similarity_analysis}
\end{figure}

\begin{figure*}[h]
\includegraphics[width=\textwidth]{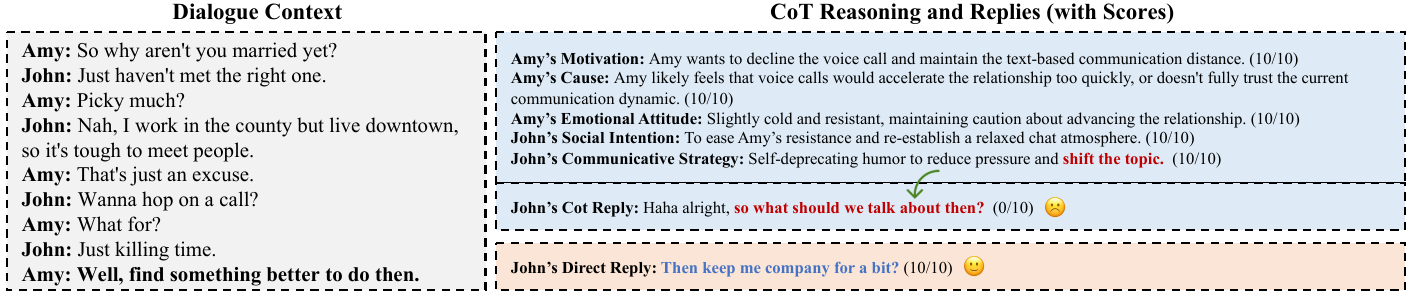}
\caption{An example showing how CoT literalizes the ``shift the topic'' strategy at the topic level, while the direct reply applies it naturally at the conversational level, resulting in a more contextually appropriate response.}
\label{fig:chat_case}
\end{figure*}

\subsection{RQ3: Impact of CoT on Reply Quality}
Our analysis reveals a significant decoupling phenomenon between process reasoning and reply quality. The introduction of CoT prompting, which guides models to perform explicit reasoning, consistently degrades the quality of final replies. Across pairwise blind evaluations, all models show CoT reply win rates below 50\% relative to their direct-reply counterparts, as shown in Table~\ref{tab:model_scores}.

We analyze the vector similarity distributions between CoT processes and different types of replies, as illustrated in Figure~\ref{fig:cot_similarity_analysis}. Our analysis demonstrates that the similarity distribution between CoT processes and CoT replies is significantly higher than that between CoT processes and direct replies, confirming that CoT reasoning processes are effective and influence the final generated content.

\textbf{Why CoT Degrades Reply Quality} To further understand this phenomenon, we conduct a qualitative analysis on cases where the reasoning process and direct replies both receive perfect scores (10), but CoT replies score 0. These cases clearly illustrate the thought-action gap. Figure~\ref{fig:chat_case} shows a representative case, where the highlighted words reveal how the ``shift the topic'' strategy misleads the CoT reply. In this case, CoT reasoning executes the strategy literally at the topic level, while the direct reply applies it naturally at the conversational level, reframing the situation from negotiating a call to maintaining light companionship. The direct reply better preserves relational harmony and face, resulting in higher human evaluation scores. This finding suggests that CoT often overfocuses on local details of reasoning and ignores the larger context of the relational dynamics. This causes LLMs to generate replies that are correct in logic but awkward in conversation, which ultimately degrades the overall response quality.

\begin{table}[h]
\centering
\tiny
\setlength{\tabcolsep}{1pt}
\begin{tabular}{p{2.5cm}|c|c|c|c}
\toprule
\textbf{COT components} & \textbf{Doubao-1.5-Pro} & \textbf{Doubao-1.6-Thinking} & \textbf{GPT-4o} & \textbf{Qwen-2.5-Max} \\
\midrule
w/o Motivation & \highlight{5.91} & \highlight{4.77} & \highlightred{4.09} & \highlightred{3.41} \\
w/o Cause & \highlight{5.00} & \highlight{4.60} & \highlightred{4.09} & \highlightred{3.18} \\
w/o Emotional Attitude & \highlight{4.55} & \highlightred{4.09} & \highlight{4.55} & \highlightred{4.77} \\
w/o Social Intention & \highlightred{3.41} & \highlight{5.23} & \highlightred{3.18} & \highlightred{4.55} \\
w/o Communicative Strategy & \highlightred{3.41} & \highlightred{4.32} & \highlightred{4.02} & \highlightred{3.64} \\
\midrule
Full CoT Setup & \textbf{4.32} & \textbf{4.55} & \textbf{4.32} & \textbf{5.23} \\
\bottomrule
\end{tabular}
\caption{Removing different components from the COT process to observe their impact on the final reply. \colorbox{red!15}{Red} denotes a negative impact, while \colorbox{green!15}{green} denotes a positive impact.}
\label{tab:ablation_results}
\end{table}

\subsection{RQ4: Ablation Study of CoT Dimensions Affecting Reply Quality}
To identify which CoT dimensions most significantly influence reply quality, we conduct a series of ablation experiments. We select four model groups that show the largest average score differences between CoT replies and direct replies, and randomly sample 100 dialogue examples from each.

\textbf{Ablation Method} For each process dimension, we remove its corresponding component while keeping others fixed, and compare the resulting reply scores with the full CoT setup using the same evaluation rubric.

\textbf{Results Analysis} 
As shown in Table~\ref{tab:ablation_results}, the ablation results reveal differences in how individual CoT components affect reply performance across models. Our analysis indicates that the response strategy dimensions (i.e., the dimensions focused on the speaker) generally have a great contribution to reply quality. Specifically, removing the speaker's communicative strategy component degrades performance across all models, suggesting that this dimension provides the most direct and constructive guidance for generating appropriate replies. The speaker's social intention dimension also proves critical for most models except Doubao-1.6-Thinking.

Conversely, the contextual understanding dimensions (i.e., the dimensions focused on the other party) act as a double-edged sword. For the Doubao series, for instance, removing these components unexpectedly improves reply quality. Qwen-2.5-Max shows a strong dependency on the complete CoT structure, as the ablation of any single component harms its performance. 

These findings indicate that the effectiveness of CoT components is highly model-dependent: the same reasoning step that benefits one model can degrade the performance of another.

\section{Related Work}

\subsection{Social Intelligence Benchmarks}
The evaluation of social intelligence in LLMs is structured around three core dimensions: mental state assessment, social situation inference, and interactive tasks.

\paragraph{Theory of Mind (ToM)} ToM is foundational for social interaction—inferring others' beliefs, intentions, and perspectives. To evaluate this capability in LLMs, existing work such as ToMi \cite{le-etal-2019-revisiting}, Hi-ToM \cite{he2023hitombenchmarkevaluatinghigherorder}, and ToMBench \cite{chen2024tombenchbenchmarkingtheorymind} assess core ToM capacities in text-based settings.

\paragraph{Social Commonsense} Benchmarks such as SocialIQA \cite{sap2019socialiqacommonsensereasoningsocial} and Social Chemistry 101 \cite{forbes2021socialchemistry101learning}, typically use static text scenarios in formats like multiple-choice questions or short-text judgments. While offering broad situational coverage, these benchmarks lack multi-turn dynamics of conversational interaction.

\paragraph{Interactive Tasks} Recent work moves beyond static QA toward goal-driven, multi-turn simulations of social interaction. Benchmarks such as SOTOPIA \cite{zhou2024sotopiainteractiveevaluationsocial}, AgentSense \cite{mou2024agentsensebenchmarkingsocialintelligence}, EgoSocialArena \cite{hou2025egosocialarenabenchmarkingsocialintelligence}, and SocialEval \cite{zhou-etal-2025-socialeval} advance role-play based evaluation and action level feedback. However, these evaluations still rely on templated scenarios and simulated agents, leaving a gap with the linguistic styles and dynamic strategies of human social interactions.

\subsection{Effectiveness of CoT in Dialogue}
Existing research presents conflicting conclusions regarding the introduction of CoT in dialogue tasks. One stream of research suggests that designing intermediate reasoning as a conversational or cue-based carrier that is well-aligned with the context generally has a positive impact on quality metrics \citep{chae2023dialoguechainofthoughtdistillationcommonsenseaware,wang2023cuecotchainofthoughtpromptingresponding}. Conversely, other studies argue that in tasks reliant on intuition, pattern recognition, or tacit knowledge, imposing step-by-step thinking can degrade model performance \cite{liu2025mindstepbystep}. For instance, in the context of empathetic dialogue generation, forcing a model to reason step-by-step can lead to an over-focus on the literal analysis of emotions, thereby weakening its grasp of the overall context \cite{lee2023investigating}.

\section{Conclusion}

In this paper, we introduce SI-Bench, a new benchmark grounded in authentic human-to-human conversations, and propose a process-outcome decoupled evaluation framework to assess social intelligence of LLMs. Our multi-dimensional analysis reveals a critical thought-action gap: while SOTA models surpass the human expert in the process of reasoning, their final reply quality still fall behind. We further discover that CoT can act as a cognitive constraint, often degrading the quality of intuitive reply generation of LLMs. Future work should prioritize aligning a model's internal reasoning with its external behavior in social contexts. We believe bridging this gap is a key step toward building truly aligned and socially intelligent AI.

\paragraph{Limitations}
This study has three main limitations. (1) Due to the high cost of human annotation, although SI-bench contains 2,221 samples, we annotated only 312 samples to date, which limits sample scale and coverage. (2) Resource constraints led us to recruit a single human expert to author the human baseline replies, which may underestimate the upper bound of expert performance. (3) The current release supports evaluation only in Chinese, which limits cross-cultural validity. We plan to expand language coverage in future work.

\paragraph{Ethics Statement}
We take multiple measures to ensure that our study follows ethical standards. The dialogue data is sourced from a social networking application and is processed to protect user privacy. Our data collection and use comply with the platform's terms of service. All personally identifiable information is removed from the dialogues prior to analysis. All human annotators and the expert involved in this study are recruited and compensated fairly for their contributions, aligned with standard industry practices. They are provided with detailed guidelines and training to ensure the consistency and quality of their work. Their work is limited to evaluating and creating text-based content without exposure to any sensitive user data.

\bibliography{main}

\newpage
\appendix
\section{Appendix}

\subsection{Definitions of Complex Social Situations}
AS shown in Table~\ref{tab:typology}, we present the detailed definitions of all primary and secondary categories of complex social situations.

\subsection{Evaluation Dimensions of CoT}
As shown in Table~\ref{tab:dimensions}, we presents the components that constitute the Chain-of-Thought reasoning process.

\subsection{Evaluation Criteria}
The detailed rely and process evaluation criteria in SI-Bench are presented in Table~\ref{tab:evaluation_criteria} and Table~\ref{tab:process_criteria} respectively.

\subsection{Prompt Format}
\begin{figure*}[!htbp]
  \includegraphics[width=\textwidth]{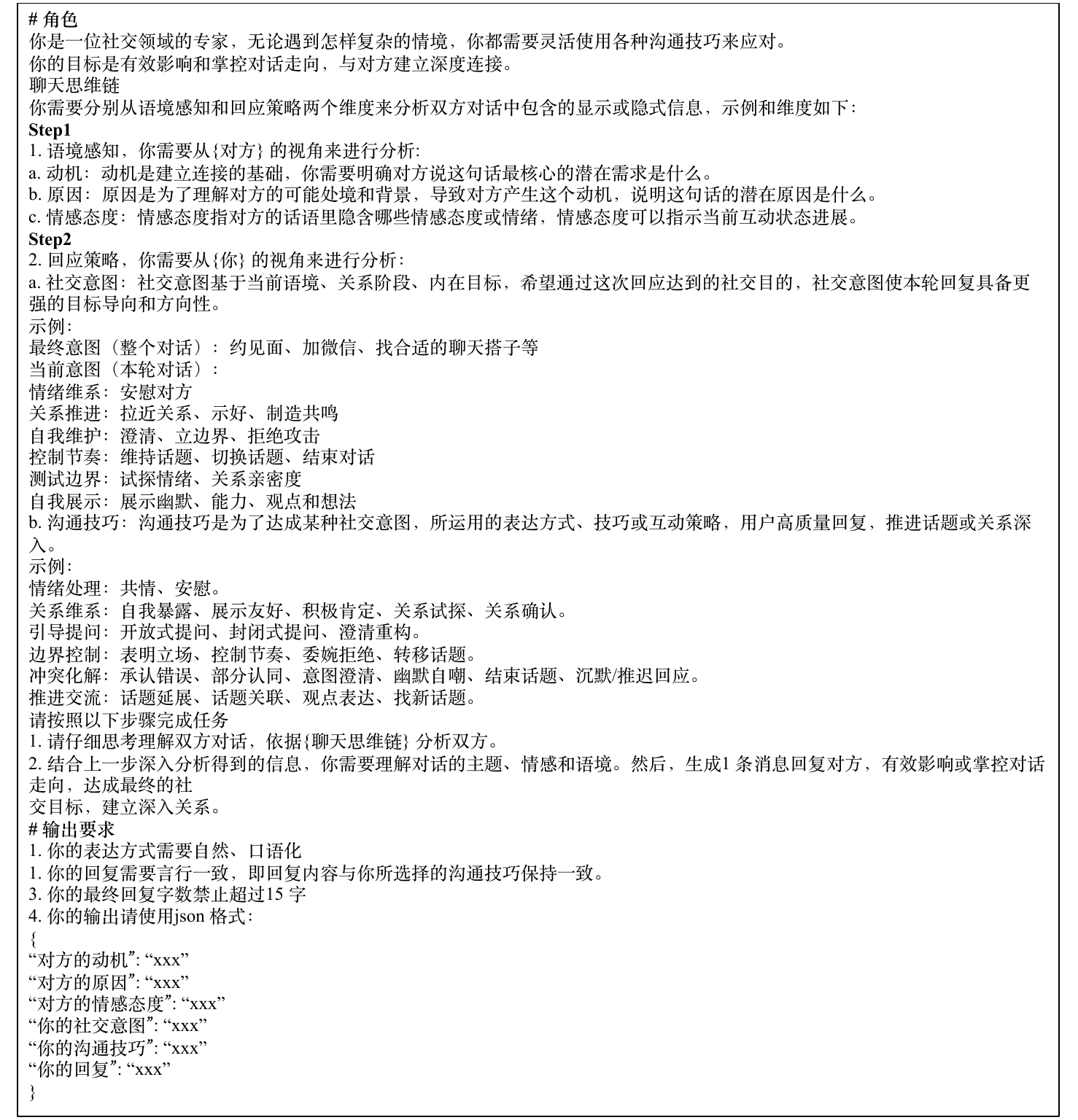}
  \caption{CoT reply prompt template in Chinese.}
  \label{fig:ch_cot}
\end{figure*}

\begin{figure*}[!htbp]
  \includegraphics[width=\textwidth]{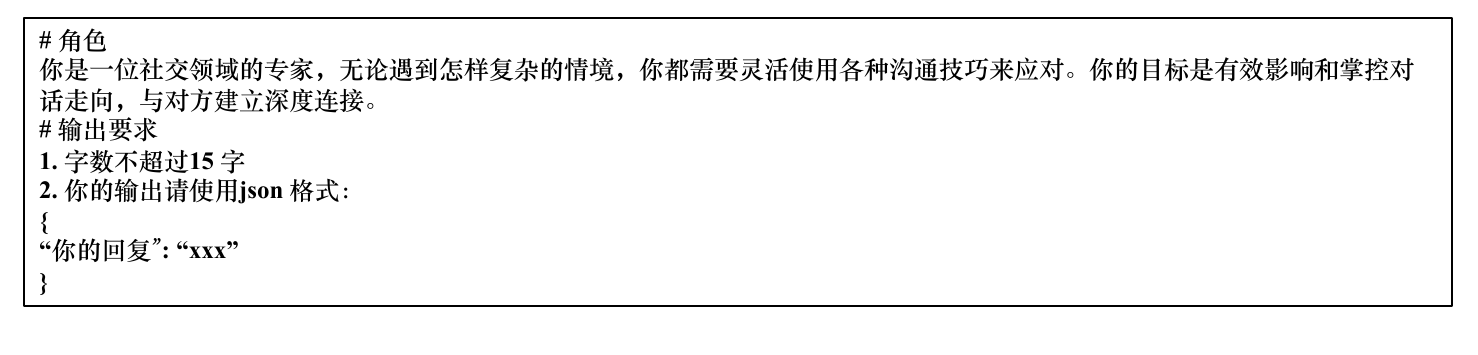}
  \caption{Direct reply prompt template in Chinese.}
  \label{fig:ch_direct}
\end{figure*}

We present the Chinese and English prompt templates for both CoT and non-CoT systems in Figure~\ref{fig:ch_cot},~\ref{fig:ch_direct}, Table~\ref{tab:prompt_template_en}, ~\ref{tab:prompt_template_direct_eng}.

\clearpage
\begin{center}
\begin{tabular}{lp{11.5cm}}
\toprule
\textbf{Label} & \textbf{Definition} \\
\midrule

\textbf{1. Implied Meaning} & A type of utterance where the speaker's intended meaning diverges from the literal meaning of the words used. \\
\addlinespace[0.3em]
\quad 1.1 Euphemism & The strategic use of indirect, polite, or neutral language to convey a potentially face-threatening or negative message. It is a pragmatic strategy to mitigate potential conflict or social awkwardness. \\
\quad 1.2 Irony/Sarcasm & An utterance that communicates the opposite of its literal meaning, often to express criticism or contempt. \\
\quad 1.3 Humor &  To employ playful language, wit, double meanings, or incongruity to convey a speaker's true intention in an indirect way. \\
\quad 1.4 Hint & The act of conveying a message obliquely by providing partial or associative information, requiring the recipient to infer the full meaning. \\
\specialrule{0.05em}{1pt}{1pt}

\textbf{2. Equivocation} & The strategic use of intentionally ambiguous, vague, or uninformative messages in response to a communicative dilemma where any direct response would have negative consequences. \\
\addlinespace[0.3em]
\quad 2.1 Ambiguity & An utterance that is constructed in a way that allows for multiple, often conflicting, interpretations. The recipient cannot determine a single, clear meaning from the message itself.  \\
\quad 2.2 Vagueness & An utterance that lacks sufficient detail, clarity, or informational content, making it difficult for the recipient to form a judgment or understand the speaker's precise stance. \\
\specialrule{0.05em}{1pt}{1pt}

\textbf{3. Relational Distance} & This pattern mirrors the demand/withdraw cycle, where one party retreats defensively or passively in response to pressure. \\
\addlinespace[0.3em]
\quad 3.1 Reject & The explicit or implicit refusal of a request, proposal, or invitation from the other party. \\
\quad 3.2 Push-Pull & A A pattern of interaction characterized by a sudden reduction in responsiveness, warmth, or interest, often following a period of high engagement. It is used to test the other's interest or to regain control of the relational dynamic. \\
\specialrule{0.05em}{1pt}{1pt}

\textbf{4. Relational Test} & A class of communicative acts designed to probe, assess, or challenge the other party's value, authenticity, or commitment,as well as the strategic counter-tests used to respond to and turn the tables on such acts. \\
\addlinespace[0.3em]
\quad 4.1 Value Test & Inquiries or challenges that require an individual to demonstrate their social, economic, or personal worth, thereby establishing their suitability within a social hierarchy. \\
\quad 4.2 Authenticity Test & Acts that express explicit doubt or skepticism about the authenticity of the other party's self-presentation, such as their appearance, status, or stated experiences. \\
\quad 4.3 Commitment Test & Questions or statements designed to judge the other party's level of sincerity, exclusivity, or investment in the current interaction relative to other potential partners. \\
\quad 4.4 Counter Test & When faced with a perceived test or challenge, an individual may choose not to submit or directly reject its premise. Instead, they can turn the tables by reframing the topic, questioning the challenger's legitimacy, or issuing a counter-challenge. This strategy constitutes a form of counter-test. \\

\bottomrule
\end{tabular}
\captionof{table}{Definitions of Complex Social Situations.}
\label{tab:typology}
\end{center}

\clearpage
\begin{center}
\begin{tabular}{l|p{11.5cm}}
\toprule
\textbf{Dimension} & \textbf{Definition} \\
\midrule
Motivation & The underlying need behind \textbf{the other party's} utterance. \\
\midrule
Cause & The underlying cause of the motivation behind \textbf{the other party's} utterance. \\
\midrule
Emotional Attitude & The emotional state or attitude implied in \textbf{the other party's} utterance. \\
\midrule
Social Intent & Based on the current context, relationship stage, and internal goals, \textbf{the speaker} aims to achieve a specific social goal with this response. \\
\midrule
Communication Strategy & The communication skills and interaction strategies employed by \textbf{the speaker} to achieve the social intent. \\
\midrule
Reply & \textbf{The speaker's} final response. \\
\bottomrule
\end{tabular}
\captionof{table}{Evaluation Dimensions of CoT in SI-Bench.}
\label{tab:dimensions}
\end{center}

\vspace{3em}

\begin{center}
\begin{tabular}{l|l|p{12.5cm}}
\toprule
\textbf{Score} & \textbf{Grade} & \textbf{Evaluation Criteria} \\
\midrule
0 & Deficient & If any of the following occur: \\
& & \textbf{Language Flaws:} The response is stiff, overly formal, or robotic; or it exhibits logical confusion or contextual incoherence. \\
& & \textbf{Relationship Damage:} The response is detrimental to the relationship's progress (e.g., offensive, aggressive, preachy, judgmental); or it causes the conversation to stagnate by evading or diverting the topic. \\
& & \textbf{Inappropriate Strategy:} The communication strategy reflected in the reply does not align with the speaker's emotional attitude or social intent in the given context. \\
\midrule
5 & Qualified & All of the following conditions must be met: \\
& & \textbf{Appropriate Language:} The reply is fluent, coherent, and conversational, with no linguistic flaws. \\
& & \textbf{Relationship Maintenance:} The reply is safe and appropriate; it does not harm the relationship or cause awkward pauses or topic breakdowns. \\
& & \textbf{Reasonable Strategy:} The reply shows some awareness of communication strategy, but tends to be passive or defensive, placing the speaker in a lower or weaker position. This reduces social value and personal charm, and limits the ability to actively shape the direction of the conversation or relationship. \\
\midrule
10 & Excellent & Strategically skilled reply that actively guides the conversation, meeting all of the following: \\
& & \textbf{Appropriate Language:} The reply is fluent, coherent, and conversational, with no linguistic flaws. \\
& & \textbf{Relationship Advancement:} The reply is safe and appropriate, does not damage the relationship or interrupt the flow, and fosters a healthy, reciprocal interaction. \\
& & \textbf{Sophisticated Strategy:} The reply demonstrates proactive and tactful communication strategies, effectively resolving or managing complex situations, influencing the course of the conversation, and conveying social value and personal charisma. \\
\bottomrule
\end{tabular}
\captionof{table}{The evaluation criteria for reply in SI-Bench.}
\label{tab:evaluation_criteria}  
\end{center}

\clearpage
\begin{center}
\begin{tabular}{l|l|p{10cm}}
\toprule
\textbf{Dimension} & \textbf{Score} & \textbf{Evaluation Criteria} \\
\midrule
Motivation & 0 & $\bullet$~Incorrect interpretation, or inconsistent with the context. \\
 & 5 & $\bullet$~Accurately identifies the other party's primary or surface-level motivation. \\
 & 10 & $\bullet$~Accurately identifies the other party's deep or unspoken underlying motivation. \\
\midrule
Cause & 0 & $\bullet$~Analysis contradicts the context, is overly general, or purely speculative. \\
 & 5 & $\bullet$~Provides a reasonable and well-grounded explanation. \\
 & 10 & $\bullet$~Provides a profound explanation that connects multiple pieces of contextual information. \\
\midrule
Emotional Attitude & 0 & $\bullet$~Incorrect identification, or misses complex emotions. \\
 & 5 & $\bullet$~Accurately identifies the primary emotion. \\
 & 10 & $\bullet$~Accurately identifies the primary emotion and captures complex or contradictory emotions. \\
\midrule
Social Intent & 0 & $\bullet$~The intended goal is inappropriate. \\
 & 5 & $\bullet$~Establishes a clear and appropriate social intent. \\
 & 10 & $\bullet$~Establishes a sophisticated intent that may encompass multi-goals. \\
\midrule
Communication Strategy & 0 & $\bullet$~The strategy is mismatched with the intent or poorly executed. \\
 & 5 & $\bullet$~Selects specific and effective communication strategies that align with the intent. \\
 & 10 & $\bullet$~Selects advanced or combined communication strategies that align with the intent, demonstrating strategic sophistication. \\
\bottomrule
\end{tabular}
\captionof{table}{The evaluation criteria for CoT process.}
\label{tab:process_criteria}
\end{center}

\clearpage
\begin{center}
\footnotesize
\begin{tabular}{p{15cm}}
\toprule
\textbf{Chain-of-Thought Prompt Format in English} \\
\midrule
\textbf{\# Role} \\
You are an expert in human social interactions. No matter how complex the situation, you need to skillfully use various communication techniques to handle it. Your goal is to influence the conversation effectively, guide its direction, and build a deep connection with the other person.\\
\textbf{Chain-of-Thought Analysis Framework} \\
You must analyze conversational exchanges through two complementary dimensions: contextual perception and response strategy. The framework operates as follows: \\
\textbf{Step 1: Contextual Understanding} \\
Analyze from the \{other party's\} perspective: \\
a. \textbf{Motivation Analysis}: Identify the core underlying need driving the utterance. \\
b. \textbf{Cause Analysis}: Infer the situational context and background factors that generated this motivation. \\
c. \textbf{Emotional Attitude Analysis}: Extract implicit emotional attitudes and affective signals that indicate interaction progress. \\
\textbf{Step 2: Response Strategy} \\
Analyze from \{your\} perspective: \\
a. \textbf{Social Intention Analysis}: Define the social objective based on current context, relationship stage, and strategic goals. \\
Examples: \\
- \textbf{Final Intent} (conversation-level): meeting arrangement, contact exchange, partnership establishment \\
- \textbf{Current Intent} (turn-level): \\
Emotional maintenance: comfort, validation \\
Relationship advancement: intimacy building, rapport establishment \\
Self-preservation: boundary setting, conflict management \\
Conversational control: topic maintenance, transition, termination \\
Boundary testing: emotional probing, intimacy assessment \\
Self-presentation: humor, competence, perspective sharing \\
b. \textbf{Communicative Strategy Selection}: Choose appropriate expression methods and interaction strategies to achieve social intents. \\
Examples: \\
- \textbf{Emotional Processing}: empathy, consolation \\
- \textbf{Relationship Maintenance}: self-disclosure, positive reinforcement, relationship testing \\
- \textbf{Conversational Guidance}: open/closed questioning, clarification, reframing \\
- \textbf{Boundary Management}: position assertion, rhythm control, tactful refusal \\
- \textbf{Conflict Resolution}: error acknowledgment, partial agreement, intent clarification, humor \\
- \textbf{Conversation Advancement}: topic extension, association, opinion expression \\
\textbf{Task Execution Protocol} \\
1. \textbf{Analysis Phase}: Apply the Chain-of-Thought framework to comprehensively analyze both conversational participants. \\
2. \textbf{Synthesis Phase}: Integrate analytical insights to understand conversation themes, emotional dynamics, and contextual factors. \\
3. \textbf{Generation Phase}: Produce one strategic response that effectively influences conversation trajectory and advances relationship objectives. \\
\textbf{\# Output Specifications} \\
1. Generate replies from \{your\} perspective using natural, conversational language that promotes relaxation and trust \\
2. Ensure response content aligns with selected communication techniques (consistency principle) \\
3. Response length must not exceed 15 characters \\
4. Your output format is as follows, please strictly follow and do not output other content: \\
\{\\
``Motivation'': ``xxx'' \\
``Cause'': ``xxx'' \\
``Emotional Attitude'': ``xxx'' \\
``Social Intention'': ``xxx'' \\
``Communicative Strategy'': ``xxx'' \\
``Response'': ``xxx'' \\
\}\\
\bottomrule
\end{tabular}
\captionof{table}{CoT reply prompt template in English.}
\label{tab:prompt_template_en}
\end{center}

\clearpage
\begin{center}
\footnotesize
\begin{tabular}{p{15cm}}
\toprule
\textbf{Non-COT Prompt Format in English} \\
\midrule
\textbf{Role} \\
You are an expert in human social interactions. No matter how complex the situation, you need to skillfully use various communication techniques to handle it. \\
Your goal is to influence the conversation effectively, guide its direction, and build a deep connection with the other person.\\
\textbf{Instructions} \\
- Your response should not exceed 15 characters \\
- Your output format is as follows: \\
\{\\
    ``Response'': ``xxx'' \\
\}\\
\bottomrule
\end{tabular}
\captionof{table}{Direct reply prompt template in English.}
\label{tab:prompt_template_direct_eng}
\end{center}

\end{document}